\documentclass[11pt,a4paper]{article}
\usepackage[hyperref]{acl2021}
\usepackage{times}
\usepackage{latexsym}

\usepackage{xurl}

\usepackage{microtype}
\usepackage{booktabs}
\usepackage{graphicx}
\usepackage{subfig}

\aclfinalcopy

\newcommand{\mf}[1]{{\color{black} {#1}}}

\title{Knowledge Distillation for Quality Estimation}

\author{
 Amit Gajbhiye,\textsuperscript{1} Marina Fomicheva,\textsuperscript{1} Fernando Alva-Manchego,\textsuperscript{1} Fr\'{e}d\'{e}ric Blain,\textsuperscript{1,2}\\
 {\bf Abiola Obamuyide},\textsuperscript{1} {\bf Nikolaos Aletras},\textsuperscript{1} {\bf Lucia Specia}\textsuperscript{1,3}\\
 \textsuperscript{1}University of Sheffield, 
 \textsuperscript{2}University of Wolverhampton, 
 \textsuperscript{3}Imperial College London \\
  \texttt{\{a.gajbhiye,m.fomicheva,f.alva,a.obamuyide,n.aletras\}@shef.ac.uk}\\
  \texttt{f.blain@wlv.ac.uk}\\
  \texttt{l.specia@imperial.ac.uk}\\
}

\date{}

\begin{document}
\maketitle
\begin{abstract}

Quality Estimation (QE) is the task of automatically predicting Machine Translation quality in the absence of reference translations, 
making it applicable in real-time settings, such as translating online social media conversations. Recent success in QE stems from the use of multilingual pre-trained representations, where very large models lead to impressive results. However, the inference time, disk and memory requirements of such models do not allow for wide usage in the real world. Models trained on distilled pre-trained  representations remain prohibitively large for many usage scenarios.
We instead propose to directly transfer knowledge from a strong QE teacher model to a much smaller model with a different, shallower architecture. We show that this approach, in combination with  data augmentation, leads to light-weight QE models that perform competitively with distilled pre-trained representations with 8x fewer parameters.

\end{abstract}

\section{Introduction}

Quality Estimation (QE) aims to predict the quality of the output of Machine Translation (MT) systems when no gold-standard  translations are available. It can make MT useful in real-world applications by informing end-users on the translation quality. 
We focus on sentence-level QE, usually formulated as a regression task where quality is required to be predicted on an 
continuous scale, e.g. 0-100.

The high performances achieved in the most recent shared task on sentence-level QE \cite{specia-etal-2020-findings-wmt} have been attributed to the use of strong pre-trained language models, namely BERT~\cite{devlin2018bert} and its multilingual variants, especially XLM-Roberta~\cite{conneau2019unsupervised}. These models have an extremely large number of parameters and, since they are required at training and inference time, they are very disk and RAM-hungry, also making inference slow. This poses challenges for real-time inference, and prohibits deployment on client machines with limited resources. 

Making models based on pre-trained representations smaller and more usable in practice is an active 
area of research. 
One approach is Knowledge Distillation (KD), aiming
to extract knowledge from a top-performing large model (the \textit{teacher}) into a smaller (in terms of memory print, computational power and prediction latency) yet well-performing model (the \textit{student}) \cite{kd2015Hinton,kd2020survey}. 
KD techniques have been used to make BERT and similar models smaller. For example, DistilBERT \cite{sanh2019distilbert} and TinyBERT \cite{tinyBERT2020Jiao} follow the same general architecture as the teacher BERT, but with a reduced number of layers. However, these  student models  are also based on Transformers and, as such, they still have too large memory and disk footprints. For instance, the number of parameters in the multilingual DistilBERT-based TransQuest model for QE \cite{ranasinghe2020transquest} is 135M.

In this paper, we propose {\bf to distill the QE model directly}, where the student architecture can be completely different from that of the teacher. Namely, we distill a large and powerful QE model based on XLM-Roberta into a small RNN-based  model.
%
Existing work along these lines has applied KD mainly to classification tasks \cite{distillingTaskSpecificBERT2019Tang,patientKD2019Sun}.
We instead explore this approach in the context of {\bf regression}. In contrast to classification, where KD provides useful information on the output distribution of incorrect classes, for regression the teacher predictions are point-based estimates, 
and as such have the same properties as gold labels. Therefore, it is not obvious whether teacher-student learning can be beneficial.  
The few existing works on KD for regression  \cite{objDectKG2017Chen,kdRegression2020Takamoto} use the teacher loss to minimise the impact of noise in the teacher predictions on the student training. However, this approach requires access to gold labelled examples to train the student, which in our case are very limited in number.

Our approach allows for much larger {\bf unlabelled student training datasets}, built only from source-MT pairs and labelled by the teacher model. 
We study the performance of student models under different training data regimes: standard training with gold labels, training with teacher predictions on the same data, training with teacher predictions on augmented in-domain and out-of-domain data, as well as augmented data filtered based on uncertainty of teacher predictions. Interestingly, we find that (i) training with teacher predictions results in better performance than training with gold labels; and (ii) student models trained with augmented data perform competitively with DistilBERT-based TransQuest predictors with 8x fewer parameters.

\section{Approach}
\label{sec:approach}

\begin{figure}
\centering
    \includegraphics[width=.45\textwidth]{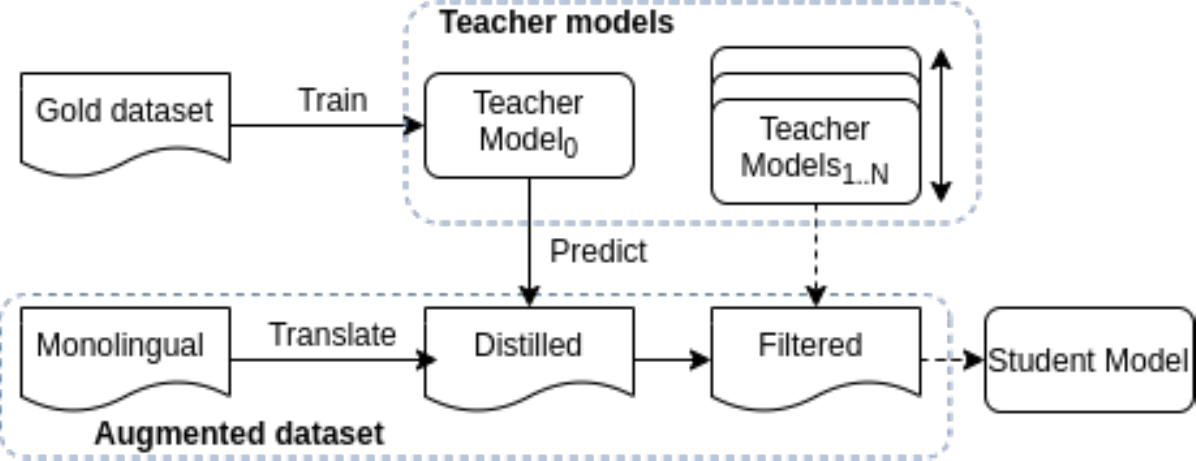}%
\caption{KD with data augmentation and noise filtering based on teacher uncertainty.}
\label{fig:data_augmentation}
\end{figure}

Figure \ref{fig:data_augmentation} summarises our approach, with the following main components:

\paragraph{Teacher-student training.} 
We use predictions from a SoTA QE model to train a light-weight student with a different architecture. Specifically, as the teacher model we use the recently proposed TransQuest QE system \cite{ranasinghe2020transquest} that fine-tunes multilingual pre-trained representations from XLM-Roberta-Large \cite{conneau2019unsupervised} to predict a continuous sentence-level quality score. For the student model, we rely on the BiRNN QE architecture proposed by \citet{iveetal-deepquest2018}.\footnote{The implementation of our student models is available at \url{https://github.com/sheffieldnlp/deepQuest-py}.}
The BiRNN model 
encodes both source and translation sentences independently using two bi-directional Recurrent Neural Networks (RNNs). The two resulting sentence representations are concatenated as the weighted sum of their word vectors, generated by an attention mechanism. For predictions at sentence-level, the weighted representation of the two input sentences is passed through a dense layer with sigmoid activation to generate the quality estimates. Table \ref{tab:efficiency} shows the number of parameters, memory and disk space requirements, as well as inference speed for the teacher model (TQ$_{\mathrm{XLM-R-Large}}$), student model (BiRNN) and TransQuest system built on DistilBERT (TQ$_{\mathrm{DistilBERT}}$). We refer the reader to Appendix \ref{app:architecture} for the details on the architecture and implementation for these models.


\begin{table}[tb]
    \centering \small
    \begin{tabular}{@{}lrp{.8cm}p{1cm}p{.8cm}@{}}
    \toprule
            &           &  \multicolumn{2}{c}{Inference} \\
    \cmidrule(lr){3-4}
    Name    & \#params & Speed (secs.) & RAM (MiB) & Disk (M) \\
    \midrule
TQ$_{\mathrm{XLM-R-Large}}$  & 561M  & 0.82 & 9,263.5  & 2140 \\
TQ$_{\mathrm{DistilBERT}}$   & 135M  & 1.09 & 1,979.2  & ~~517 \\
BiRNN                        &  18M  & 0.39 & ~~155.6 & ~~132 \\
    
    \bottomrule
    \end{tabular}
    \caption{Efficiency. Inference speed and RAM for prediction are for 1 sentence on CPU (Intel Xeon Silver 4114 CPU @ 2.20GHz).}
    \label{tab:efficiency}
\end{table}

In classification tasks, KD benefits learning as it uses information on the output distribution and has an effect akin to label smoothing \cite{tang2020understanding}. In regression, teacher labels are instead point-wise estimates just like the gold labels. 
Existing work on KD for regression uses teacher loss to minimise the impact of noise in the teacher predictions on student training \cite{kdRegression2020Takamoto}. However, this approach is not suitable for QE as we have access to a very limited number of gold-labelled examples. We propose a simple strategy that relies directly on teacher predictions for training the student model. 

\paragraph{Data augmentation.} The power of QE models based on pre-trained representations is due to the rich knowledge that comes from training Transformer-based language models on very large amounts of data. Typically, much smaller datasets are available for downstream tasks, which suffice for fine-tuning but that are hardly suitable for training a neural model from scratch. We exploit the teacher-student framework to produce additional training data. Specifically, we first generate MT outputs for a set of sentences in the source language and domain of interest using the same MT system that was used for generating the test data. Second, we use the teacher model described above 
to produce predictions. These predictions are then used as labels for training the student. 

\paragraph{Noise filtering.} The benefits of data augmentation can be hampered by noise in teacher predictions. In a classification setting, where the student loss is computed with respect to the output distribution of the teacher model, this issue is ameliorated by the example re-weighting effect where teacher predictions with higher confidence have an overall higher impact on learning \cite{furlanello2018born}. Previous work has used teacher loss to address this issue for regression \cite{objDectKG2017Chen}. However, this strategy is not suitable for data augmentation as it requires both gold labels and teacher predictions. 

As an alternative, we propose a mechanism to filter-out noisy examples in the augmented dataset based on uncertainty quantification. Recent work has shown that ensembles produce accurate uncertainty estimates \cite{lakshminarayanan2017simple}. We exploit this idea by training a set of additional teacher models independently on the same training data using random initialisation, and using the variance of their predictions as an indicator of predictive uncertainty.\footnote{Here we use ensemble only as a way of estimating the error in the predictions and leave distillation based on ensemble predictions to future work.} Intuitively, examples with very high variance would correspond to noisy teacher predictions. We filter out from the student training data the instances where the variance is more than one standard deviation away from its mean value. This is expected to have a higher impact on the results in the out-of-domain setting where the performance of the teacher model is less stable and teacher predictions can contain more noise. 



\section{Experiments}
\label{sec:experiments}

\paragraph{MLQE Dataset.} For training the teacher and for evaluation, we use the MLQE dataset 
~\cite{tacl2020}, same as in the WMT2020 QE Shared Task \cite{specia-etal-2020-findings-wmt}. 
This dataset contains sentences extracted from Wikipedia translated to and from English for a total of six language pairs: English--German (En-De),\footnote{We skip this language pair as the performance of the teacher model for it is too weak.} English--Chinese (En-Zh), Romanian--English (Ro-En), Estonian--English (Et-En), Sinhala--English (Si-En) and Nepali--English (Ne-En).
Each translation was produced with a SoTA Transformer-based NMT model and manually annotated for quality using an annotation scheme inspired by the Direct Assessment methodology \citep{graham-EtAl:2013:LAW7-ID}. The scores are produced on a continuous scale indicating perceived translation quality in 0-100. For each language pair, this dataset contains partitions for
training (7K), dev (1K), and test (1K).

\paragraph{Distilled dataset.} Monolingual data for data augmentation was sampled from Wikipedia following the procedure described in \citet{tacl2020} to preserve the domain of the MLQE dataset. Specifically, we sampled documents from Wikipedia for English, Estonian, Romanian, Sinhalese and Nepalese and selected the top 100 documents containing the largest number of
sentences that are: (i) in the intended source language according to a language-id classifier and (ii) have the length between 50 and 150 characters. Table \ref{tab:data} shows the total amount of sentences in the monolingual Wikipedia dataset collected for data augmentation.

To test the impact of data domain on the performance of the student QE models, we also collect out-of-domain data for the Et-En language pair. The out-of-domain data is sampled from Common Crawl. We use the version of Common Crawl distributed by the WMT2018 News Translation Task\footnote{ \url{http://www.statmt.org/wmt18/translation-task.html}}. The total amount of sentences in this dataset is 100,779,314.

\begin{table}[t]
\centering
\begin{tabular}{@{}lr@{}}
\toprule
Language & Sentences \\
\midrule
Estonian & 25,176 \\
Romanian & 372,690 \\
Sinhala & 139,406 \\
Nepalese & 85,343 \\
English & 1,563,519 \\
\bottomrule
\end{tabular}
\caption{Number of sentences extracted from Wikipedia for data augmentation.}
\label{tab:data}
\end{table}

\begin{table*}[t!]
    \centering \small
    \begin{tabular}{@{}llccccc@{}}
    \toprule
    Name    & Training data & Et-En & Ro-En & Si-En & Ne-En & En-Zh \\
    \midrule
    TQ$\mathrm{_{TEACHER}}$ & MLQE-gold & 0.77 & 0.88 & 0.60 & 0.75 & 0.44 \\
\cmidrule{1-7}
BiRNN$\mathrm{_{STUDENT}}$ & MLQE-dist & 0.45 & 0.62 & 0.44 & 0.46 & {\bf 0.18} \\
BiRNN$\mathrm{_{STUDENT+AUG}}$ & Wiki-dist & {\bf 0.50} & {\bf 0.69} & {\bf 0.45} & {\bf 0.54} & { 0.17} \\
\cmidrule{1-7}
\cmidrule{1-7}
BiRNN & MLQE-gold & 0.37 & 0.60 & 0.40 & 0.42 & 0.15 \\
Predictor-Estimator & MLQE-gold & 0.48 & 0.69 & 0.37 & 0.39 & 0.19 \\
TQ$\mathrm{_{DistilBERT}}$ & MLQE-gold & 0.62 & 0.78 & 0.51 & 0.61 & 0.36 \\
    \bottomrule
    \end{tabular}
    \caption{Pearson correlation with human judgments on the MLQE test set. MLQE-gold:  training partition of MLQE dataset; MLQE-dist: distilled version of the MLQE training set with teacher predictions  used as labels;  Wiki-dist: the Wikipedia dataset produced by data augmentation. 
    Boldface results indicate our best student models.}
    \label{tab:distillation}
\end{table*}

To translate the data, we used the same MT models that generated the test data, built with fairseq \cite{ott2019fairseq} and made available by the WMT2020 QE Shared Task organisers.\footnote{\url{https://github.com/facebookresearch/mlqe/tree/master/nmt_models}.} Sentences that were part of the training data for the MT models or part of the MLQE dataset were excluded. We generate quality predictions for the remaining sentences using the teacher models, as described in Section \ref{sec:approach}. We used a random subset of 100K sentences from Wikipedia to train the student model for each of the language pairs except for Et-En where the total amount of collected in-domain monolingual data is 25K. 

\paragraph{Models.} As teachers, we use pre-trained models from TransQuest (TQ$_{\mathrm{TEACHER}}$),  
one of the winning submissions in the WMT2020 QE Shared Task, which we fine-tuned on the MLQE dataset. 
For noise filtering, we train five teacher models with random initialisation. 
As students, we use BiRNN models from DeepQuest \cite{iveetal-deepquest2018}. 
We also compare our results against the Predictor-Estimator model \citep{kim2017predictor,openkiwi}, the baseline at the WMT2020 QE Shared Task, and TransQuest models using multilingual DistilBERT.\footnote{Multilingual DistilBERT is available at \url{ https://huggingface.co/distilbert-base-multilingual-cased}. We follow the same training procedure as for the teacher model described in detail in Appendix \ref{app:architecture}.}

\section{Results} 


Table \ref{tab:distillation} shows the Pearson correlation with human judgments on the test partition of the MLQE dataset for different models and specifies the type of data used for training.\footnote{TQ$_{\mathrm{TEACHER}}$, TQ$\mathrm{_{DistilBERT}}$ and Predictor-Estimator use contextual representations trained on large amounts of additional data, which are then fine-tuned for the QE task.} The correlation for the student models (BiRNN$\mathrm{_{STUDENT*}}$) does not reach the performance of TQ$_{\mathrm{TEACHER}}$. Smaller models may lack representation power for modeling cross-lingual tasks such as QE. Also, distillation for regression is more challenging, as discussed in Section \ref{sec:approach}. However, \textbf{training on the in-domain distilled data} (BiRNN$\mathrm{_{STUDENT+AUG}}$) \textbf{allows to obtain performances comparable to DistilBERT} (TQ$\mathrm{_{DistilBERT}}$) \textbf{with much lighter models} (see Table \ref{tab:efficiency}).\footnote{This is true for all language pairs except Et-En and En-Zh. For Et-En we have a considerably smaller amount of in-domain data available for training, whereas for En-Zh the teacher model appears to be too weak to be useful for KD.} Furthermore, this approach results in a substantial improvement over shallow models trained on gold data (BiRNN and Predictor-Estimator) for all of the language pairs. The student performance for each language pair is strongly related to the performance of the teacher. Thus, the Ro-En student achieves the highest correlation results, whereas correlation for En-Zh is weak.

\begin{table}[t]
    \centering \small
    \begin{tabular}{@{}lcccc@{}}
    \toprule
     & Ro-En & Si-En & Ne-En & En-Zh \\
    \midrule
10K & 0.56 ±.00 & 0.36 ±.00 & 0.41 ±.00 & 0.09 ±.01 \\
50K & 0.64 ±.00 & 0.45 ±.01 & 0.53 ±.00 & 0.20 ±.03 \\
70K & 0.66 ±.00 & 0.46 ±.01 & 0.54 ±.00 & 0.19 ±.02 \\
100K & 0.69 ±.00 & 0.47 ±.02 & 0.54 ±.00 & 0.17 ±.02 \\
\bottomrule
\end{tabular}
\caption{Pearson correlation on the test partition of the MLQE dataset for BiRNN student models trained with different amounts of distilled Wikipedia data.}
\label{tab:distillBySize}
\end{table}

\mf{We further analyse what is the impact of different data selection strategies on the results.} \mf{\textbf{First}, we sample random subsets of training instances from the Wikipedia distilled dataset and evaluate the performance of the student model trained with this data. We run the training 3 times with different random splits for training and validation and report the mean and confidence intervals. Table \ref{tab:distillBySize} shows the results for all languages where we have enough Wikipedia data (for Et-En we only have 25K in total). The largest boost in correlation is observed when going from 10K to 50K.}

\begin{figure}
\centering
    \includegraphics[width=.48\textwidth]{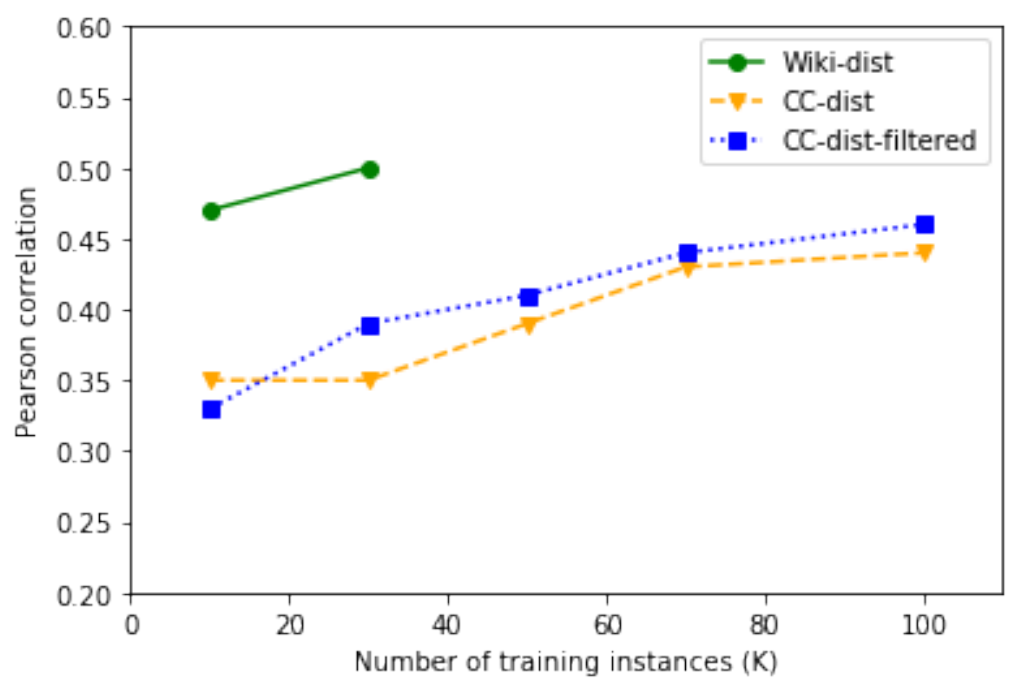}
\caption{Pearson correlation results on the MLQE test set for the student models trained with different amounts of distilled data in-domain (Wiki-dist), out-of-domain (CC-dist) and out-of-domain with noise filtering (CC-dist-filtered), for Et-En.}
\label{fig:performance_by_domain}
\end{figure}

\textbf{Second}, we compare  student models trained on these subsets of  distilled data of different sizes, i.e. using data extracted from Wikipedia (in-domain), against data splits of the same size extracted from Common Crawl (out-of-domain). For the out-of-domain data we apply the noise filtering strategy described in Section \ref{sec:approach}. Figure \ref{fig:performance_by_domain} shows the results for Et-En, where our largest in-domain set has 25K sentences. We observe that using in-domain data appears to be much more effective than sampling larger amounts of generic data. Noise filtering gives some improvement in the results but its effect appears to be marginal compared to the effect of training with in-domain data.

\begin{figure}
\centering
    \includegraphics[width=.45\textwidth]{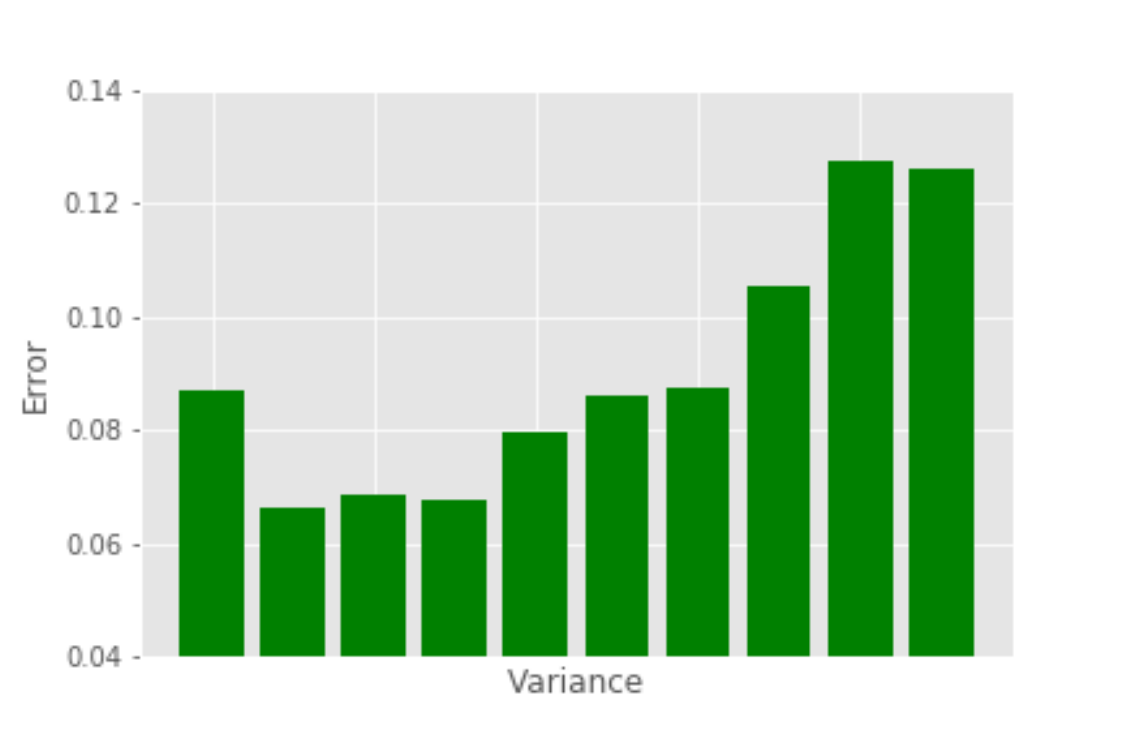}
\caption{Variance in the predictions of 6 teacher models trained with different random seed against predictions error on the test partition of Ro-En MLQE dataset.}
\label{fig:teacher_variance_test20}
\end{figure}

Figure \ref{fig:teacher_variance_test20} provides an illustration of the relation between the variance in the predictions of multiple teacher models and prediction error for Ro-En language pair on the in-domain data. We group the sentences in the test partitions of MLQE dataset in 10 bins according to the variance between the predictions of the different teacher models in the ensemble. We then calculate the average prediction error in each bin, where the error is the absolute difference between model predictions and human judgements. As shown in Figure \ref{fig:teacher_variance_test20}, higher variance in the predictions indeed corresponds to larger prediction error.

\begin{figure}
\centering
    \includegraphics[width=.4\textwidth]{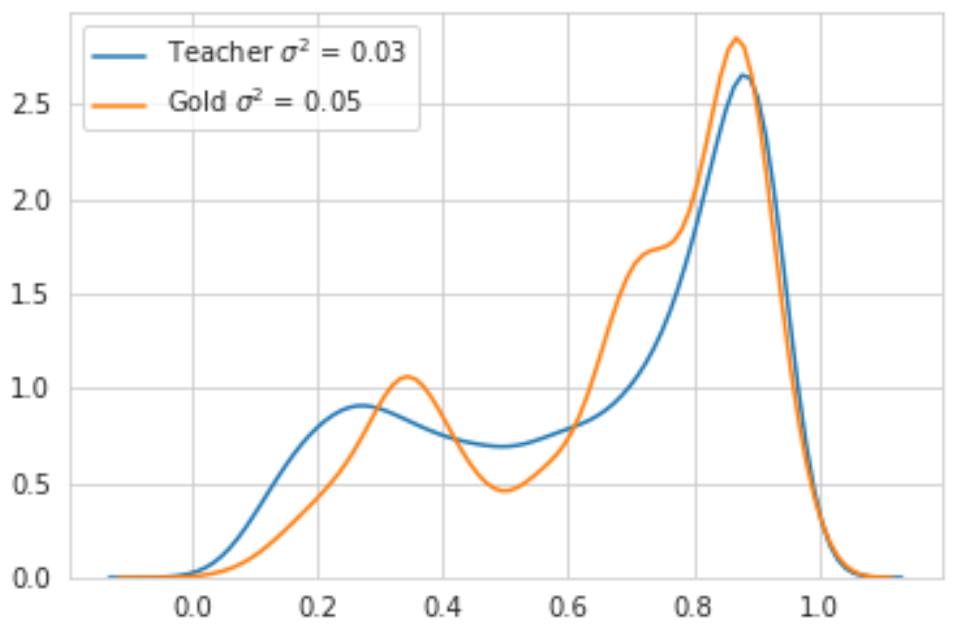}
\caption{Distribution of teacher scores (blue) and gold labels (orange) on the training partitions of Et-En MLQE dataset.}
\label{fig:eten_teacher_scores}
\end{figure}

Interestingly, from Table \ref{tab:distillation} we see that  \textbf{training with distilled data brings benefits even without data augmentation} for some of the language pairs. The correlation for Et-En improves from 0.37 to 0.45 by training on teacher predictions (BiRNN$_{\mathrm{STUDENT}}$) instead of gold labels (BiRNN) on the same MLQE dataset. To gain an intuition for this improvement, Figure \ref{fig:eten_teacher_scores} shows the distribution of teacher predictions and human scores on the train partition of MLQE dataset. We hypothesize that teacher predictions having a smoother distribution with reduced variance makes learning easier. As shown in Appendix \ref{app:distribution}, we observe this trend for all language pairs in the dataset.

\section{Conclusions}


In this paper, we showed that knowledge distillation, through a teacher-student approach that directly distills QE predictions, can be effective in building a light-weight QE model with  similar performance to a SoTA architecture trained on distilled yet large pre-trained representations. We also introduced a noise filtering approach that leverages the uncertainty of an ensemble of teacher models to determine which training instances should be discarded when training the student models, which can be beneficial especially for data augmentation from out-of-domain sources.
This results in QE models 4x smaller in disk space with 8x fewer parameters, and 3x faster in inference speed.


\section*{Acknowledgements}
This work was supported by funding from the Bergamot project (EU H2020 Grant No. 825303).

\bibliographystyle{acl_natbib}
\bibliography{references.bib}

\begin{thebibliography}{23}
\expandafter\ifx\csname natexlab\endcsname\relax\def\natexlab#1{#1}\fi

\bibitem[{Chen et~al.(2017)Chen, Choi, Yu, Han, and
  Chandraker}]{objDectKG2017Chen}
Guobin Chen, Wongun Choi, Xiang Yu, Tony~X. Han, and Manmohan Chandraker. 2017.
\newblock \href
  {https://proceedings.neurips.cc/paper/2017/hash/e1e32e235eee1f970470a3a6658dfdd5-Abstract.html}
  {Learning efficient object detection models with knowledge distillation}.
\newblock In \emph{Advances in Neural Information Processing Systems 30: Annual
  Conference on Neural Information Processing Systems 2017, December 4-9, 2017,
  Long Beach, CA, {USA}}, pages 742--751.

\bibitem[{Conneau et~al.(2020{\natexlab{a}})Conneau, Khandelwal, Goyal,
  Chaudhary, Wenzek, Guzm{\'a}n, Grave, Ott, Zettlemoyer, and
  Stoyanov}]{conneau2019unsupervised}
Alexis Conneau, Kartikay Khandelwal, Naman Goyal, Vishrav Chaudhary, Guillaume
  Wenzek, Francisco Guzm{\'a}n, Edouard Grave, Myle Ott, Luke Zettlemoyer, and
  Veselin Stoyanov. 2020{\natexlab{a}}.
\newblock \href {https://doi.org/10.18653/v1/2020.acl-main.747} {Unsupervised
  cross-lingual representation learning at scale}.
\newblock In \emph{Proceedings of the 58th Annual Meeting of the Association
  for Computational Linguistics}, pages 8440--8451.

\bibitem[{Conneau et~al.(2020{\natexlab{b}})Conneau, Khandelwal, Goyal,
  Chaudhary, Wenzek, Guzm{\'{a}}n, Grave, Ott, Zettlemoyer, and
  Stoyanov}]{xmlr2020Conneau}
Alexis Conneau, Kartikay Khandelwal, Naman Goyal, Vishrav Chaudhary, Guillaume
  Wenzek, Francisco Guzm{\'{a}}n, Edouard Grave, Myle Ott, Luke Zettlemoyer,
  and Veselin Stoyanov. 2020{\natexlab{b}}.
\newblock \href {https://doi.org/10.18653/v1/2020.acl-main.747} {Unsupervised
  cross-lingual representation learning at scale}.
\newblock In \emph{Proceedings of the 58th Annual Meeting of the Association
  for Computational Linguistics, {ACL} 2020, Online, July 5-10, 2020}, pages
  8440--8451. Association for Computational Linguistics.

\bibitem[{Devlin et~al.(2018)Devlin, Chang, Lee, and
  Toutanova}]{devlin2018bert}
Jacob Devlin, Ming-Wei Chang, Kenton Lee, and Kristina Toutanova. 2018.
\newblock \href {https://arxiv.org/abs/1810.04805} {Bert: Pre-training of deep
  bidirectional transformers for language understanding}.
\newblock \emph{arXiv preprint arXiv:1810.04805}.

\bibitem[{Fomicheva et~al.(2020)Fomicheva, Sun, Yankovskaya, Blain, Guzmán,
  Fishel, Aletras, Chaudhary, and Specia}]{tacl2020}
Marina Fomicheva, Shuo Sun, Lisa Yankovskaya, Frédéric Blain, Francisco
  Guzmán, Mark Fishel, Nikolaos Aletras, Vishrav Chaudhary, and Lucia Specia.
  2020.
\newblock \href {https://doi.org/10.1162/tacl\_a\_00330} {Unsupervised quality
  estimation for neural machine translation}.
\newblock \emph{Transactions of the Association for Computational Linguistics},
  8:539--555.

\bibitem[{Furlanello et~al.(2018)Furlanello, Lipton, Tschannen, Itti, and
  Anandkumar}]{furlanello2018born}
Tommaso Furlanello, Zachary Lipton, Michael Tschannen, Laurent Itti, and Anima
  Anandkumar. 2018.
\newblock \href
  {https://proceedings.mlr.press/v80/furlanello18a/furlanello18a.pdf} {Born
  again neural networks}.
\newblock In \emph{International Conference on Machine Learning}, pages
  1607--1616. PMLR.

\bibitem[{Gou et~al.(2020)Gou, Yu, Maybank, and Tao}]{kd2020survey}
Jianping Gou, Baosheng Yu, Stephen~John Maybank, and Dacheng Tao. 2020.
\newblock \href {https://link.springer.com/article/10.1007/s11263-021-01453-z}
  {Knowledge distillation: A survey}.
\newblock \emph{arXiv preprint arXiv:2006.05525}.

\bibitem[{Graham et~al.(2013)Graham, Baldwin, Moffat, and
  Zobel}]{graham-EtAl:2013:LAW7-ID}
Yvette Graham, Timothy Baldwin, Alistair Moffat, and Justin Zobel. 2013.
\newblock \href {https://www.aclweb.org/anthology/W13-2305.pdf} {Continuous
  measurement scales in human evaluation of machine translation}.
\newblock In \emph{Proceedings of the 7th Linguistic Annotation Workshop and
  Interoperability with Discourse}, pages 33--41.

\bibitem[{Hinton et~al.(2015)Hinton, Vinyals, and Dean}]{kd2015Hinton}
Geoffrey~E. Hinton, Oriol Vinyals, and Jeffrey Dean. 2015.
\newblock \href {http://arxiv.org/abs/1503.02531} {Distilling the knowledge in
  a neural network}.
\newblock \emph{CoRR}, abs/1503.02531.

\bibitem[{Ive et~al.(2018)Ive, Blain, and Specia}]{iveetal-deepquest2018}
Julia Ive, Fr{\'e}d{\'e}ric Blain, and Lucia Specia. 2018.
\newblock \href {https://wlv.openrepository.com/handle/2436/623557}
  {Deep{Q}uest: a framework for neural-based quality estimation}.
\newblock In \emph{Proceedings of COLING 2018, the 27th International
  Conference on Computational Linguistics: Technical Papers}, Santa Fe, new
  Mexico.

\bibitem[{Jiao et~al.(2020)Jiao, Yin, Shang, Jiang, Chen, Li, Wang, and
  Liu}]{tinyBERT2020Jiao}
Xiaoqi Jiao, Yichun Yin, Lifeng Shang, Xin Jiang, Xiao Chen, Linlin Li, Fang
  Wang, and Qun Liu. 2020.
\newblock \href {https://doi.org/10.18653/v1/2020.findings-emnlp.372}
  {Tinybert: Distilling {BERT} for natural language understanding}.
\newblock In \emph{Proceedings of the 2020 Conference on Empirical Methods in
  Natural Language Processing: Findings, {EMNLP} 2020, Online Event, 16-20
  November 2020}, pages 4163--4174. Association for Computational Linguistics.

\bibitem[{Kepler et~al.(2019)Kepler, Tr\'{e}nous, Treviso, Vera, and
  Martins}]{openkiwi}
F\'{a}bio Kepler, Jonay Tr\'{e}nous, Marcos Treviso, Miguel Vera, and Andr\'{e}
  F.~T. Martins. 2019.
\newblock \href {https://www.aclweb.org/anthology/P19-3020} {Open{K}iwi: An
  open source framework for quality estimation}.
\newblock In \emph{Proceedings of the 57th Annual Meeting of the Association
  for Computational Linguistics--System Demonstrations}, pages 117--122,
  Florence, Italy. Association for Computational Linguistics.

\bibitem[{Kim et~al.(2017)Kim, Lee, and Na}]{kim2017predictor}
Hyun Kim, Jong-Hyeok Lee, and Seung-Hoon Na. 2017.
\newblock \href {https://www.aclweb.org/anthology/W17-4763/}
  {Predictor-estimator using multilevel task learning with stack propagation
  for neural quality estimation}.
\newblock In \emph{Proceedings of the Second Conference on Machine
  Translation}, pages 562--568.

\bibitem[{Lakshminarayanan et~al.(2017)Lakshminarayanan, Pritzel, and
  Blundell}]{lakshminarayanan2017simple}
Balaji Lakshminarayanan, Alexander Pritzel, and Charles Blundell. 2017.
\newblock \href {https://arxiv.org/abs/1612.01474} {{Simple and Scalable
  Predictive Uncertainty Estimation Using Deep Ensembles}}.
\newblock In \emph{Advances in Neural Information Processing Systems}, pages
  6402--6413.

\bibitem[{Ott et~al.(2019)Ott, Edunov, Baevski, Fan, Gross, Ng, Grangier, and
  Auli}]{ott2019fairseq}
Myle Ott, Sergey Edunov, Alexei Baevski, Angela Fan, Sam Gross, Nathan Ng,
  David Grangier, and Michael Auli. 2019.
\newblock \href {https://arxiv.org/abs/1904.01038} {fairseq: A fast, extensible
  toolkit for sequence modeling}.
\newblock In \emph{Proceedings of NAACL-HLT 2019: Demonstrations}.

\bibitem[{Ranasinghe et~al.(2020)Ranasinghe, Orasan, and
  Mitkov}]{ranasinghe2020transquest}
Tharindu Ranasinghe, Constantin Orasan, and Ruslan Mitkov. 2020.
\newblock \href {https://arxiv.org/abs/2011.01536} {Transquest: Translation
  quality estimation with cross-lingual transformers}.
\newblock In \emph{Proceedings of the 28th International Conference on
  Computational Linguistics}.

\bibitem[{Sanh et~al.(2019)Sanh, Debut, Chaumond, and
  Wolf}]{sanh2019distilbert}
Victor Sanh, Lysandre Debut, Julien Chaumond, and Thomas Wolf. 2019.
\newblock \href {https://arxiv.org/abs/1910.01108} {Distilbert, a distilled
  version of bert: smaller, faster, cheaper and lighter}.
\newblock \emph{arXiv preprint arXiv:1910.01108}.

\bibitem[{Specia et~al.(2020)Specia, Blain, Fomicheva, Fonseca, Chaudhary,
  Guzm{\'a}n, and Martins}]{specia-etal-2020-findings-wmt}
Lucia Specia, Fr{\'e}d{\'e}ric Blain, Marina Fomicheva, Erick Fonseca, Vishrav
  Chaudhary, Francisco Guzm{\'a}n, and Andr{\'e} F.~T. Martins. 2020.
\newblock \href {https://www.aclweb.org/anthology/2020.wmt-1.79} {Findings of
  the {WMT} 2020 shared task on quality estimation}.
\newblock In \emph{Proceedings of the Fifth Conference on Machine Translation},
  pages 743--764, Online. Association for Computational Linguistics.

\bibitem[{Sun et~al.(2019)Sun, Cheng, Gan, and Liu}]{patientKD2019Sun}
Siqi Sun, Yu~Cheng, Zhe Gan, and Jingjing Liu. 2019.
\newblock \href {https://doi.org/10.18653/v1/D19-1441} {Patient knowledge
  distillation for {BERT} model compression}.
\newblock In \emph{Proceedings of the 2019 Conference on Empirical Methods in
  Natural Language Processing and the 9th International Joint Conference on
  Natural Language Processing, {EMNLP-IJCNLP} 2019, Hong Kong, China, November
  3-7, 2019}, pages 4322--4331. Association for Computational Linguistics.

\bibitem[{Takamoto et~al.(2020)Takamoto, Morishita, and
  Imaoka}]{kdRegression2020Takamoto}
Makoto Takamoto, Yusuke Morishita, and Hitoshi Imaoka. 2020.
\newblock \href {https://doi.org/10.1109/MIPR49039.2020.00021} {An efficient
  method of training small models for regression problems with knowledge
  distillation}.
\newblock In \emph{3rd {IEEE} Conference on Multimedia Information Processing
  and Retrieval, {MIPR} 2020, Shenzhen, China, August 6-8, 2020}, pages 67--72.
  {IEEE}.

\bibitem[{Tang et~al.(2020)Tang, Shivanna, Zhao, Lin, Singh, Chi, and
  Jain}]{tang2020understanding}
Jiaxi Tang, Rakesh Shivanna, Zhe Zhao, Dong Lin, Anima Singh, Ed~H Chi, and
  Sagar Jain. 2020.
\newblock \href {https://arxiv.org/abs/2002.03532} {Understanding and improving
  knowledge distillation}.
\newblock \emph{arXiv preprint arXiv:2002.03532}.

\bibitem[{Tang et~al.(2019)Tang, Lu, Liu, Mou, Vechtomova, and
  Lin}]{distillingTaskSpecificBERT2019Tang}
Raphael Tang, Yao Lu, Linqing Liu, Lili Mou, Olga Vechtomova, and Jimmy Lin.
  2019.
\newblock \href {http://arxiv.org/abs/1903.12136} {Distilling task-specific
  knowledge from {BERT} into simple neural networks}.
\newblock \emph{CoRR}, abs/1903.12136.

\bibitem[{Vaswani et~al.(2017)Vaswani, Shazeer, Parmar, Uszkoreit, Jones,
  Gomez, Kaiser, and Polosukhin}]{vaswani2017attention}
Ashish Vaswani, Noam Shazeer, Niki Parmar, Jakob Uszkoreit, Llion Jones,
  Aidan~N Gomez, {\L}ukasz Kaiser, and Illia Polosukhin. 2017.
\newblock \href {https://arxiv.org/abs/1706.03762} {Attention is all you need}.
\newblock In \emph{Advances in Neural Information Processing Systems}, pages
  5998--6008.

\end{thebibliography}

\appendix
\clearpage

\begin{figure*}[t]%
    \centering
    \subfloat{{\includegraphics[width=0.4\textwidth]{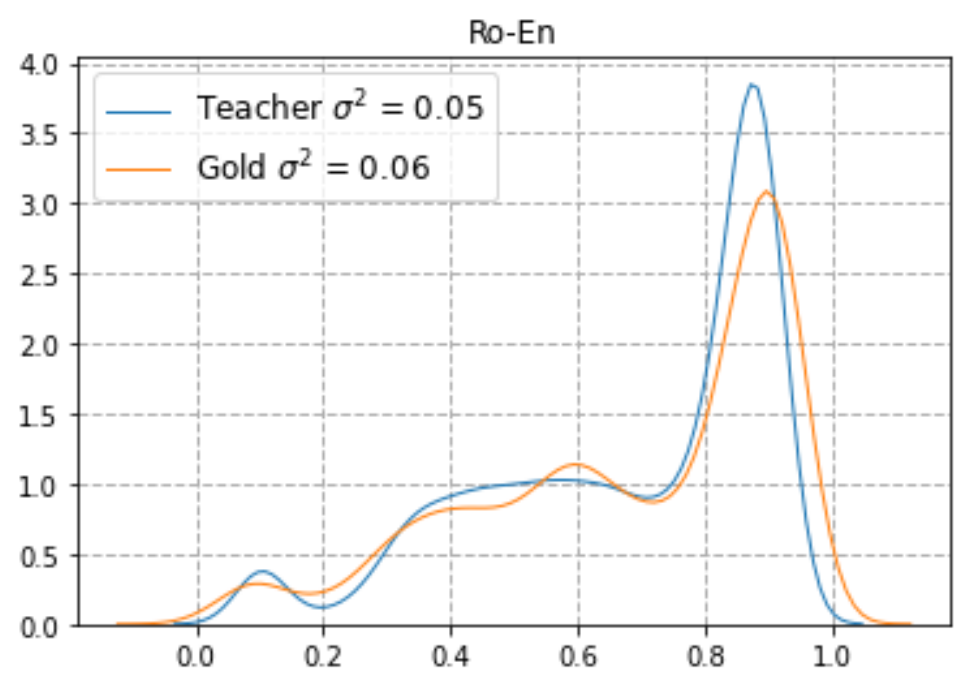} }}%
    \,
    \subfloat{{\includegraphics[width=0.4\textwidth]{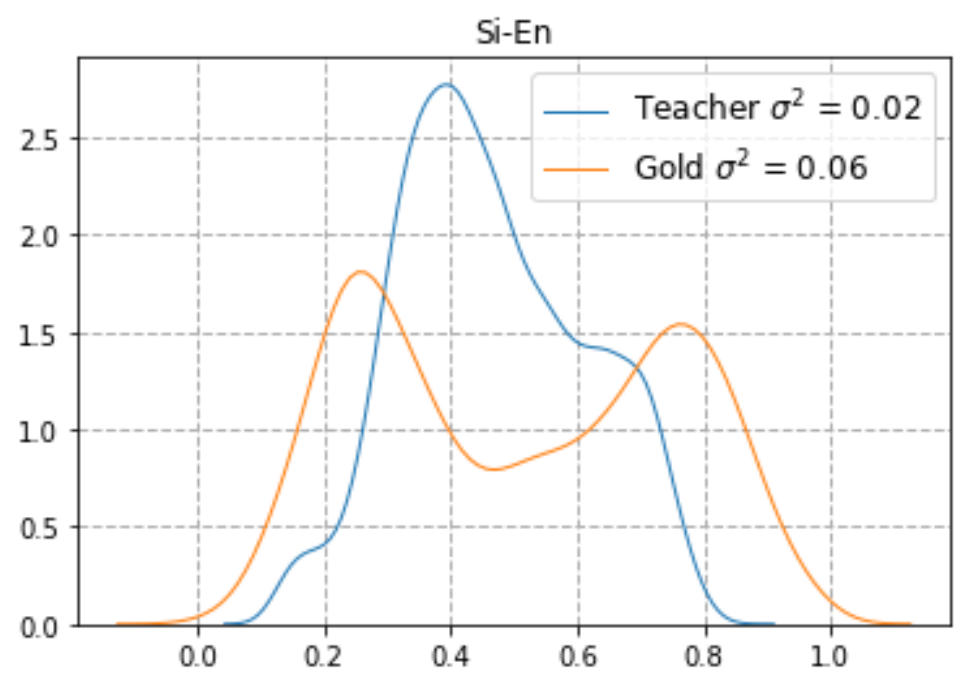}}}%
    \,
    \subfloat{{\includegraphics[width=0.4\textwidth]{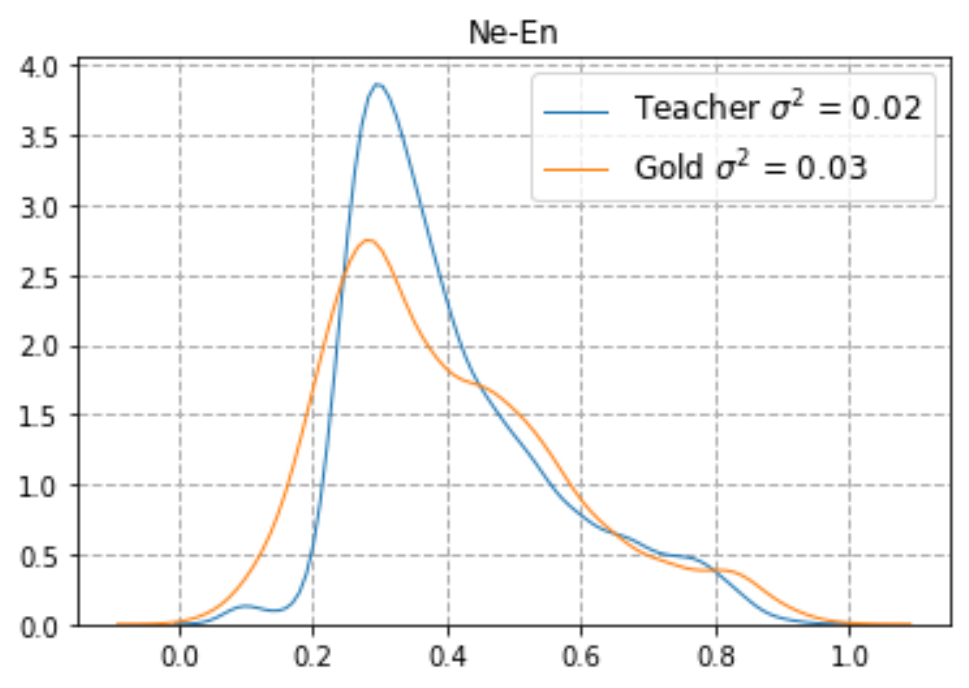}}}%
    \,
    \subfloat{{\includegraphics[width=0.4\textwidth]{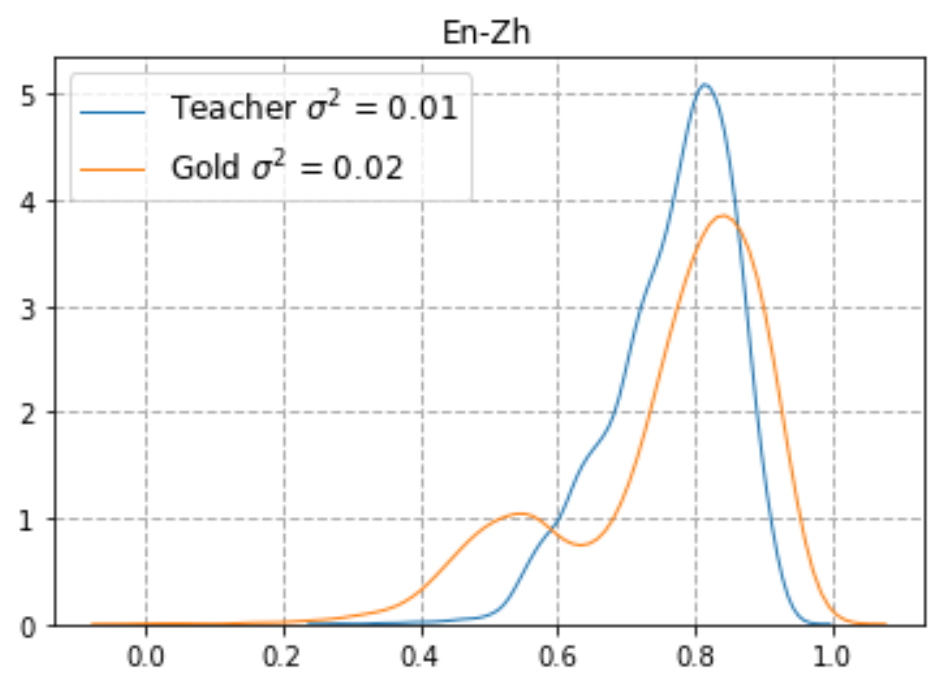} }}%
    \,
    \caption{Distribution of teacher scores (blue) and gold labels (orange) on the training partitions for different language pairs in the MLQE dataset.}%
    \label{fig:appendix_distribution_teacher}%
\end{figure*}

\section{Teacher and student models}
\label{app:architecture}

\paragraph{Teacher model} For the teacher model, we use the transformer-based MonoTransQuest \cite{ranasinghe2020transquest} architecture of the TransQuest framework with XLMR-large \cite{xmlr2020Conneau} as the underlying pre-trained representation model. XLMR is Transformer \cite{vaswani2017attention} based masked language model (with $24$ Transformer blocks and the vocabulary size of 250K) trained on one hundred languages using approximately two terabytes of CommonCrawl data. MonoTransQuest takes as input the concatenation of the original sentence and its translation separated by the the special $[SEP]$ token. The learned representation of the special $[CLS]$ token is considered as the joint representation of the original and translated sentence. The joint representation is then fed to the final $softmax$ layer to predict the quality score of the translation. For distillation we use the models that were made available for download by the authors.\footnote{\url{https://tharindudr.github.io/TransQuest/pretrained/\#available-models}} For training the additional teacher models for data filtering we follow the training settings in \citet{ranasinghe2020transquest}: we used a batch size of 8,
Adam optimiser with learning rate 2e--5, and a linear learning rate warm-up over 10\% of the training data. During the training process, the parameters of XLM-R model, as well as the parameters of the subsequent layers, are updated. All the models were trained for 3 epochs.

\paragraph{Student model}
For the student model, we rely on the BiRNN QE architecture proposed by \citet{iveetal-deepquest2018}. Our implementation of this architecture is available for download.\footnote{\url{https://github.com/sheffieldnlp/deepQuest-py}} The light-weight architecture (15 layers) of this model is as follows: both source and target sentences are independently encoded by a dedicated embedding layer followed by a bi-directional Recurrent Neural Network (RNN).
The two resulting sentence representations are then concatenated as a weighted sum of their word vectors, generated by an attention mechanism. 
The resulting representation is then passed through an output dense layer with sigmoid activation to generate the quality estimates.
We use the BiRNN model in its default configuration: both source and target embeddings are of size 300, each encoder has a hidden size of 50. The vocabulary size is limited to the 30k most common words. The model is trained with early stopping with a patience of 5.

\section{Output Distribution for Teacher Models}
\label{app:distribution}

Figure \ref{fig:appendix_distribution_teacher} shows the distribution of teacher scores (blue) and gold labels (orange) on the training partitions for the language pairs in MLQE dataset.







\end{document}